\documentclass{article}

\usepackage{times}

\usepackage{graphicx} 
\usepackage{subfigure}
\usepackage{natbib}

\usepackage{algorithm}
\usepackage{algorithmic}
\usepackage{amsmath}
\usepackage{amsfonts}
\usepackage{amssymb}
\usepackage{xcolor}

\usepackage{hyperref}

\DeclareMathOperator*{\argmin}{argmin}

\usepackage[accepted]{icml2017}

\icmltitlerunning{Language Modeling with Gated Convolutional Networks}
\title{Language Modeling with Gated Convolutional Networks}

\begin{document}

\twocolumn[
\icmltitle{Language Modeling with Gated Convolutional Networks}

\begin{icmlauthorlist}
\icmlauthor{Yann N. Dauphin}{a}
\icmlauthor{Angela Fan}{a}
\icmlauthor{Michael Auli}{a}
\icmlauthor{David Grangier}{a}
\end{icmlauthorlist}
\icmlaffiliation{a}{Facebook AI Research}

\icmlcorrespondingauthor{Yann N. Dauphin}{ynd@fb.com}

\icmlkeywords{convolutional neural nets, recurrent nets, language modeling}

\vskip 0.3in
]

\printAffiliationsAndNotice{}  


\begin{abstract}
The pre-dominant approach to language modeling to date is based on recurrent neural networks. Their success on this task is often linked to their ability to capture unbounded context. In this paper we develop a finite context approach through stacked convolutions, which can be more efficient since they allow parallelization over sequential tokens. We propose a novel simplified gating mechanism that outperforms \citet{oord2016conditional} and investigate the impact of key architectural decisions. The proposed approach achieves state-of-the-art on the WikiText-103 benchmark, even though it features long-term dependencies, as well as competitive results on the Google Billion Words benchmark. Our model reduces the latency to score a sentence by an order of magnitude compared to a recurrent baseline. To our knowledge, this is the first time a non-recurrent approach is competitive with strong recurrent models on these large scale language tasks.
\end{abstract}

\section{Introduction}

Statistical language models estimate the probability distribution of a sequence of words by modeling the probability of the next word given preceding words, i.e.
\begin{align}
P(w_0, \ldots, w_N) = P(w_0)\prod_{i=1}^N P(w_i|w_0,\ldots, w_{i-1})\nonumber,
\end{align}
where $w_i$ are discrete word indices in a vocabulary.
Language models are a critical part of systems for speech recognition \cite{YuDengASR} and machine translation \cite{KoehnSMT}.

Recently, neural networks \cite{bengio2003neural,mikolov:2010:interspeech,jozefowicz2016exploring} have been shown to outperform classical \emph{n-gram} language models \cite{kneser1995improved,chen1996empirical}.
These classical models suffer from data sparsity, which makes it difficult to represent large contexts and thus, long-range dependencies.
Neural language models tackle this issue by embedding words in continuous space over which a neural network is applied.
The current state of the art for language modeling is based on long short term memory networks (LSTM; Hochreiter et al., 1997)\nocite{hochreiter1997long} which can theoretically model arbitrarily long dependencies.

In this paper, we introduce new gated convolutional networks and apply them to language modeling.
Convolutional networks can be stacked to represent large context sizes and extract hierarchical features over larger and larger contexts with more abstractive features \cite{lecun1995convolutional}.
This allows them to model long-term dependencies by applying $\mathcal{O}(\frac{N} {k})$ operations over a context of size $N$ and kernel width $k$.
In contrast, recurrent networks view the input as a chain structure and therefore require a linear number $\mathcal{O}(N)$ of operations.

Analyzing the input hierarchically bears resemblance to classical grammar formalisms which build syntactic tree structures of increasing granuality, e.g., sentences consist of noun phrases and verb phrases each comprising further internal structure \citep{manning1999foundations,steedman2002book}.
Hierarchical structure also eases learning since the number of non-linearities for a given context size is reduced compared to a chain structure, thereby mitigating the vanishing gradient problem \citep{glorot2010understanding}.

Modern hardware is well suited to models that are highly parallelizable.
In recurrent networks, the next output depends on the previous hidden state which does not enable parallelization over the elements of a sequence. Convolutional networks, however, are very amenable to this computing paradigm since the computation of all input words can be performed simultaneously (\textsection\ref{sec:gcnn}).

Gating has been shown to be essential for recurrent neural networks to reach state-of-the-art performance \cite{jozefowicz2016exploring}.
Our gated linear units reduce the vanishing gradient problem for deep architectures by providing a linear path for the gradients while retaining non-linear capabilities (\textsection\ref{sec:gating}).


We show that gated convolutional networks outperform other recently published language models such as LSTMs trained in a similar setting on the Google Billion Word Benchmark \citep{chelba2013one}.
We also evaluate the ability of our models to deal with long-range dependencies on the WikiText-103 benchmark for which the model is conditioned on an entire paragraph rather than a single sentence and we achieve a new state-of-the-art on this dataset \citep{2016arXiv160907843M}.
Finally, we show that gated linear units achieve higher accuracy and converge faster than the LSTM-style gating of Oord et al. (2016; \textsection\ref{sec:setup}, \textsection\ref{sec:results})\nocite{oord2016conditional}.

\section{Approach}
\label{sec:gcnn}






In this paper we introduce a new neural language model that replaces the recurrent connections typically used in recurrent networks with gated temporal convolutions.
Neural language models \citep{bengio2003neural} produce a representation ${\bf H} = [{\bf h}_0, \ldots, {\bf h}_N]$ of the context for each word $w_0,\ldots, w_N$ to predict the next word $P(w_i|{\bf h}_i)$.
Recurrent neural networks $f$ compute ${\bf H}$ through a recurrent function ${\bf h}_i = f({\bf h}_{i-1}, w_{i-1})$ which is an inherently sequential process that cannot be parallelized over $i$.\footnote{Parallelization is usually done over multiple sequences instead.}

Our proposed approach convolves the inputs with a function $f$ to obtain ${\bf H} = f \ast w$ and therefore has no temporal dependencies, so it is easier to parallelize over the individual words of a sentence.
This process will compute each context as a function of a number of preceding words.
Compared to recurrent networks, the context size is finite but we will demonstrate both that infinite contexts are not necessary and our models can represent large enough contexts to perform well in practice (\textsection\ref{sec:results}).

\begin{figure}[ht]
   \centering
   \includegraphics[width=.5\textwidth]{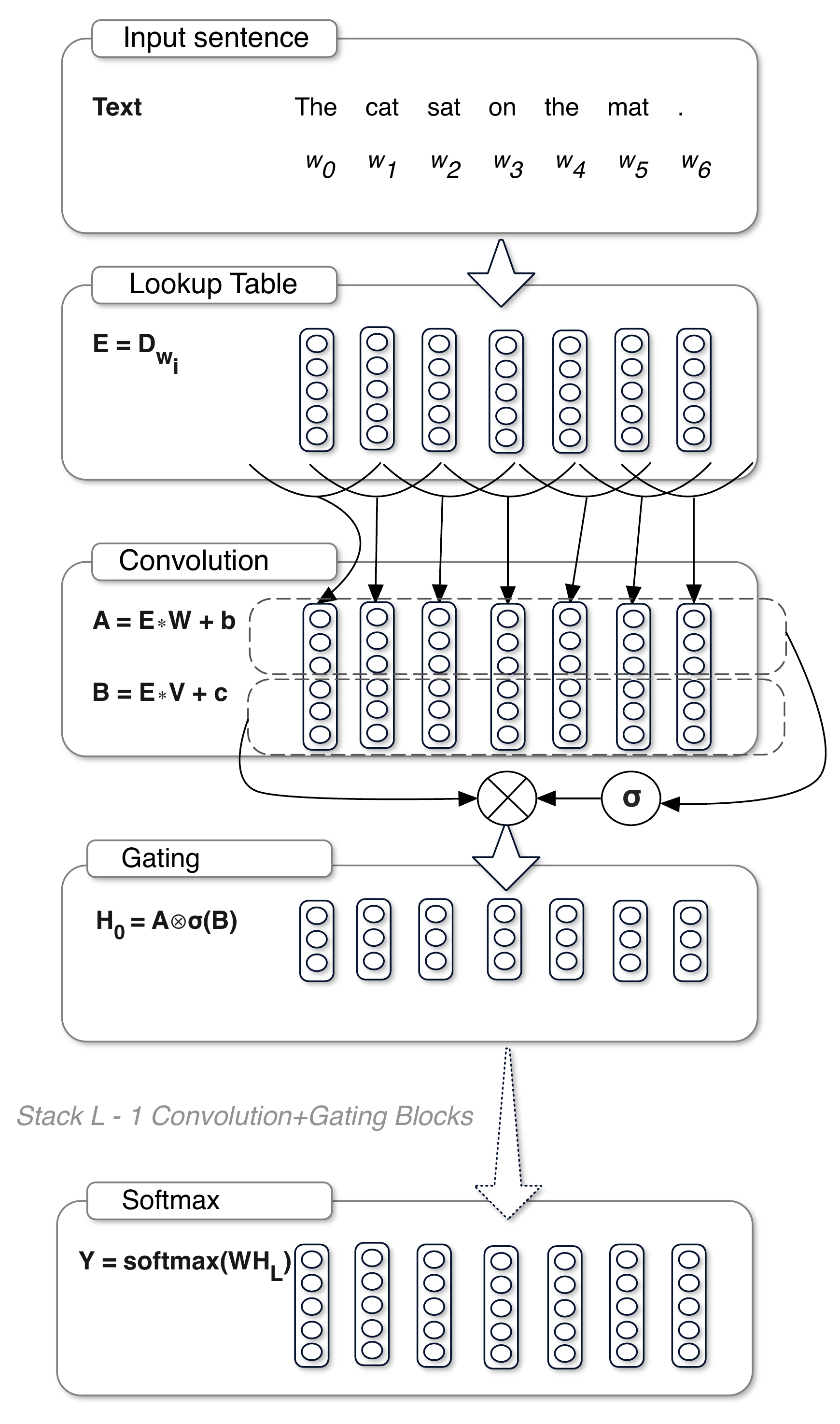}
 \caption{Architecture of the gated convolutional network for language modeling.}
 \label{fig:architecture}
\end{figure}

Figure~\ref{fig:architecture} illustrates the model architecture.
Words are represented by a vector embedding stored in a lookup table ${\bf D}^{|\mathcal{V}|\times e}$ where $|\mathcal{V}|$ is the number of words in the vocabulary and $e$ is the embedding size.
The input to our model is a sequence of words $w_0, \ldots, w_N$ which are represented by word embeddings ${\bf E} = [{\bf D}_{w_0}, \ldots, {\bf D}_{w_{N}}]$.
We compute the hidden layers $h_0, \ldots, h_L$ as
\begin{align}\label{eqn:glu}
h_l({\bf X}) = ({\bf X} \ast {\bf W} + {\bf b})\otimes \sigma({\bf X} \ast {\bf V} + {\bf c})
\end{align}
where $m,n$ are respectively the number of input and output feature maps and $k$ is the patch size, ${\bf X}\in \mathbb{R}^{N\times m}$ is the input of layer $h_l$ (either word embeddings or the outputs of previous layers), ${\bf W}\in \mathbb{R}^{k\times m \times n}$, ${\bf b}\in \mathbb{R}^n$, ${\bf V}\in \mathbb{R}^{k\times m \times n}$, ${\bf c}\in \mathbb{R}^n$ are learned parameters, $\sigma$ is the sigmoid function and $\otimes$ is the element-wise product between matrices.

When convolving inputs, we take care that ${\bf h}_i$ does not contain information from future words.
We address this by shifting the convolutional inputs to prevent the kernels from seeing future context \cite{oord2016pixel}.
Specifically, we zero-pad the beginning of the sequence with $k - 1$ elements, assuming the first input element is the beginning of sequence marker which we do not predict and $k$ is the width of the kernel.

The output of each layer is a linear projection ${\bf X} \ast {\bf W} + {\bf b}$ modulated by the gates $\sigma({\bf X} \ast {\bf V} + {\bf c})$.
Similar to LSTMs, these gates multiply each element of the matrix ${\bf X} \ast {\bf W} + {\bf b}$ and control the information passed on in the hierarchy.
We dub this gating mechanism Gated Linear Units (GLU).
Stacking multiple layers on top of the input ${\bf E}$ gives a representation of the context for each word ${\bf H} = h_L\circ\ldots\circ h_0({\bf E})$.
We wrap the convolution and the gated linear unit in a pre-activation residual block that adds the input of the block to the output \citep{he2015deep}.
The blocks have a bottleneck structure for computational efficiency and each block has up to 5 layers.

The simplest choice to obtain model predictions is to use a softmax layer,
but this choice is often computationally inefficient for large vocabularies and approximations such as noise contrastive estimation \citep{gutmann2010noise} or hierarchical softmax \citep{morin2005hierarchical} are preferred.
We choose an improvement of the latter known as \emph{adaptive softmax} which assigns higher capacity to very frequent words and lower capacity to rare words \citep{2016arXiv160904309G}.
This results in lower memory requirements as well as faster computation at both training and test time.




\section{Gating Mechanisms}
\label{sec:gating}

Gating mechanisms control the path through which information flows in the network and have proven to be
useful for recurrent neural networks \cite{hochreiter1997long}.
LSTMs enable long-term memory via a separate cell controlled by \emph{input} and \emph{forget} gates.
This allows information to flow unimpeded through potentially many timesteps.
Without these gates, information could easily vanish through the transformations of each timestep.
In contrast, convolutional networks do not suffer from the same kind of vanishing gradient and we find experimentally that they do not require forget gates.

Therefore, we consider models possessing solely output gates, which allow the network to control what information should be propagated through the hierarchy of layers.
We show this mechanism to be useful for language modeling as it allows the model to select which words or features are relevant for predicting the next word.
Parallel to our work, \citet{oord2016conditional} have shown the effectiveness of an LSTM-style mechanism of the form $\text{tanh}({\bf X} \ast {\bf W} + {\bf b})\otimes \sigma({\bf X} \ast {\bf V} + {\bf c})$ for the convolutional modeling of images. Later, \citet{kalchbrenner2016bytenet} extended this mechanism with additional gates for use in translation and character-level language modeling.

Gated linear units are a simplified gating mechanism based on the work of \citet{dauphin2015predicting} for non-deterministic gates that reduce the vanishing gradient problem by having linear units coupled to the gates.
This retains the non-linear capabilities of the layer while allowing the gradient to propagate through the linear unit without scaling.
The gradient of the LSTM-style gating of which we dub gated tanh unit (GTU) is
\begin{align}
\nabla[\text{tanh}({\bf X})\otimes \sigma({\bf X})] = \text{tanh}'({\bf X})\nabla{\bf X}\otimes \sigma({\bf X}) \nonumber\\
+ \sigma'({\bf X})\nabla {\bf X}\otimes \text{tanh}({\bf X}).
\end{align}
Notice that it gradually vanishes as we stack layers because of the downscaling factors $\text{tanh}'({\bf X})$ and $\sigma'({\bf X})$.
In contrast, the gradient of the gated linear unit
\begin{align}
\nabla[{\bf X}\otimes \sigma({\bf X})] = \nabla{\bf X}\otimes \sigma({\bf X}) + {\bf X}\otimes \sigma'({\bf X})\nabla {\bf X}
\end{align}
has a path $\nabla{\bf X}\otimes \sigma({\bf X})$ without downscaling for the activated gating units in $\sigma({\bf X})$.
This can be thought of as a multiplicative skip connection which helps gradients flow through the layers. We compare the different gating schemes experimentally in Section \textsection\ref{sec:gating} and we find gated linear units allow for faster convergence to better perplexities.

\section{Experimental Setup}
\label{sec:setup}

\subsection{Datasets}

We report results on two public large-scale language modeling datasets.
First, the Google Billion Word dataset \cite{chelba2013one} is considered one of the largest language modeling datasets with almost one billion tokens and a vocabulary of over 800K words.
In this dataset, words appearing less than 3 times are replaced with a special unknown symbol.
The data is based on an English corpus of $30,301,028$ sentences whose order has been shuffled.
Second, WikiText-103 is a smaller dataset of over 100M tokens with a vocabulary of about 200K words \citep{2016arXiv160907843M}.
Different from GBW, the sentences are consecutive which allows models to condition on larger contexts rather than single sentences.
For both datasets, we add a beginning of sequence marker $\text{\textless S \textgreater}$ at the start of each line and an end of sequence marker $\text{\textless/S\textgreater}$ at the end of each line.
On the Google Billion Word corpus each sequence is a single sentence, while on WikiText-103 a sequence is an entire paragraph.
The model sees $\text{\textless S\textgreater}$ and $\text{\textless/S \textgreater}$ as input but only predicts the end of sequence marker $\text{\textless/S\textgreater}$.
We evaluate models by computing the perplexity $\text{e}^{\frac{1}{N}\sum_i^N-\log p(w_i|\ldots, w_{i-1})}$ on the standard held out test portion of each dataset.

\renewcommand{\arraystretch}{1.2}
\begin{table*}[ht]
  \centering \small
  \begin{tabular}{l | c | c | c | c | c | c}
    \hline
    Name & GCNN-13 & GCNN-14B & GCNN-9 & GCNN-8B & GCNN-8 & GCNN-14\\
    \hline
    Dataset & \multicolumn{4}{c|}{Google Billion Word} & \multicolumn{2}{c}{wikitext-103}\\
    \hline
    Lookup & \multicolumn{4}{c|}{128} & \multicolumn{2}{c}{280}\\
    \hline
        Conv1 & $[4,1268]\times1$ & $[5,512]\times1$ & $[4,807]\times1$ & $[1,512]\times1$ & $[4,900]\times1$ & $[6, 850] \times3$ \\
    \hline
        Conv2.x & $\left[ \begin{array}{c} 4,1268 \\ 4,1268  \end{array}\right] \times 12$ & $\left[ \begin{array}{c} 1,128 \\ 5,128 \\ 1,512  \end{array}\right] \times 3$ & $\left[ \begin{array}{c} 4,807 \\ 4,807  \end{array}\right] \times 4$ & $\left[ \begin{array}{c} 1,128 \\ 5,128 \\ 1,512  \end{array}\right] \times 3$ & $\left[4,900\right]\times7$ & $[1, 850] \times1$\\
    \hline
        Conv3.x & & $\left[ \begin{array}{c} 1,512 \\ 5,512 \\ 1,1024  \end{array}\right] \times 3$ &  & $\left[ \begin{array}{c} 1,256 \\ 5,256 \\ 1,512  \end{array}\right] \times 3$ & & $[5, 850] \times4$\\
    \hline
        Conv4.x & & $\left[ \begin{array}{c} 1,1024 \\ 5,1024 \\ 1,2048  \end{array}\right] \times 6$ & & $\left[ \begin{array}{c} 1,1024 \\ 1,1024 \\ 1,2048  \end{array}\right] \times 1$ & & $[1, 850] \times1$\\
    \hline
        Conv5.x & & $\left[ \begin{array}{c} 1,1024 \\ 5,1024 \\ 1,4096  \end{array}\right]\times 1$ &  & & & $[4, 850] \times3$\\
    \hline
    Conv6.x & & & & & & $[4, 1024] \times1$\\
    \hline
    Conv7.x & & & & & & $[4, 2048] \times1$\\
    \hline
    AdaSoftmax & \multicolumn{2}{c|}{10k,40k,200k} & \multicolumn{2}{c|}{4k,40k,200k} & 2k,10k,50k & 10k,20k,200k \\
    \hline
    \end{tabular}
   \caption{Architectures for the models. The residual building blocks are shown in brackets with the format $[k,n]$. ``B'' denotes bottleneck architectures.}
 \label{fig:models}
\end{table*}

\renewcommand{\thefootnote}{\fnsymbol{footnote}}
\begin{table*}[ht]
  \centering
  \begin{tabular}{ l c r }
    \hline
    Model & Test PPL & Hardware \\
    \hline
        Sigmoid-RNN-2048 \cite{ji2015blackout} & 68.3 & 1 CPU \\
        Interpolated KN 5-Gram \cite{chelba2013one} & 67.6 & 100 CPUs \\
        Sparse Non-Negative Matrix LM \cite{shazeer2014skip} & 52.9 & - \\
        RNN-1024 + MaxEnt 9 Gram Features \cite{chelba2013one} & 51.3 & 24 GPUs \\
        LSTM-2048-512 \cite{jozefowicz2016exploring} & 43.7 & 32 GPUs \\
        2-layer LSTM-8192-1024 \cite{jozefowicz2016exploring} & 30.6 & 32 GPUs \\
        BIG GLSTM-G4 \cite{KuchaievG17} & 23.3\footnotemark[1] & 8 GPUs \\
        \hline
        LSTM-2048 \cite{2016arXiv160904309G} & 43.9 & 1 GPU \\
        2-layer LSTM-2048 \cite{2016arXiv160904309G} & 39.8 & 1 GPU \\
        GCNN-13 & 38.1 & 1 GPU \\
        GCNN-14 {\small Bottleneck} & 31.9 & 8 GPUs \\
    \end{tabular}
   \caption{Results on the Google Billion Word test set. The GCNN outperforms the LSTMs with the same output approximation.}
 \label{fig:gbw}
\end{table*}

\subsection{Training}

We implement our models in Torch \citep{collobert2011torch} and train on Tesla M40 GPUs. The majority of our models are trained on single GPU, as we focused on identifying compact architectures with good generalization and efficient computation at test time. We trained larger models with an 8-GPU setup by copying the model onto each GPU and dividing the batch such that each worker computes 1/8th of the gradients. The gradients are then summed using Nvidia NCCL. The multi-GPU setup allowed us to train models with larger hidden units. 


We train using Nesterov's momentum \citep{sutskever2013importance}.
While the cost in terms of memory is storing another vector of the size of the parameters, it increases the speed of convergence significantly with minimal additional computation compared to standard stochastic gradient descent.
The speed of convergence was further increased with gradient clipping \citep{pascanu2013difficulty} and weight normalization \citep{salimans2016weight}.

\citet{pascanu2013difficulty} argue for gradient clipping because it prevents the gradient explosion problem that characterizes RNNs.
However, gradient clipping is not tied to RNNs, as it can be derived from the general concept of trust region methods. Gradient clipping is found using a spherical trust region
\begin{align}
\Delta \theta^* & = \argmin_{%
      \substack{%
        \text{s.\,t.}\, \|\Delta \theta\| \leq \epsilon
      }
    }
    f(\theta) + \nabla f^T \Delta \theta \nonumber\\
    & = - \max(\|\nabla f\|, \epsilon) \frac{\nabla f}{\|\nabla f\|}.
\end{align}
Empirically, our experiments converge significantly faster with the use of gradient clipping even though we do not use a recurrent architecture.

In combination, these methods led to stable and fast convergence with comparatively large learning rates such as $1$.

\subsection{Hyper-parameters}

We found good hyper-parameter configurations by cross-validating with random search on a validation set.
For model architecture, we select the number of residual blocks between $\{1, \ldots, 10\}$, the size of the embeddings with $\{128, \ldots, 256\}$, the number of units between $\{128, \ldots, 2048\}$, and the kernel width between $\{3, \ldots, 5\}$.
In general, finding a good architecture was simple and the rule of thumb is that the larger the model, the better the performance.
In terms of optimization, we initialize the layers of the model with the Kaiming initialization \cite{he2015delving}, with the learning rate sampled uniformly in the interval $[1., 2.]$, the momentum set to $0.99$, and clipping set to $0.1$.
Good hyper-parameters for the optimizer are quite straightforward to find and the optimal values do not change much between datasets.

\section{Results}
\label{sec:results}

LSTMs and recurrent networks are able to capture long term dependencies and are fast becoming cornerstones in natural language processing.
In this section, we compare strong LSTM and RNN models from the literature to our gated convolutional approach on two datasets.

\begin{figure}[ht]
   \centering
   \includegraphics[width=.5\textwidth]{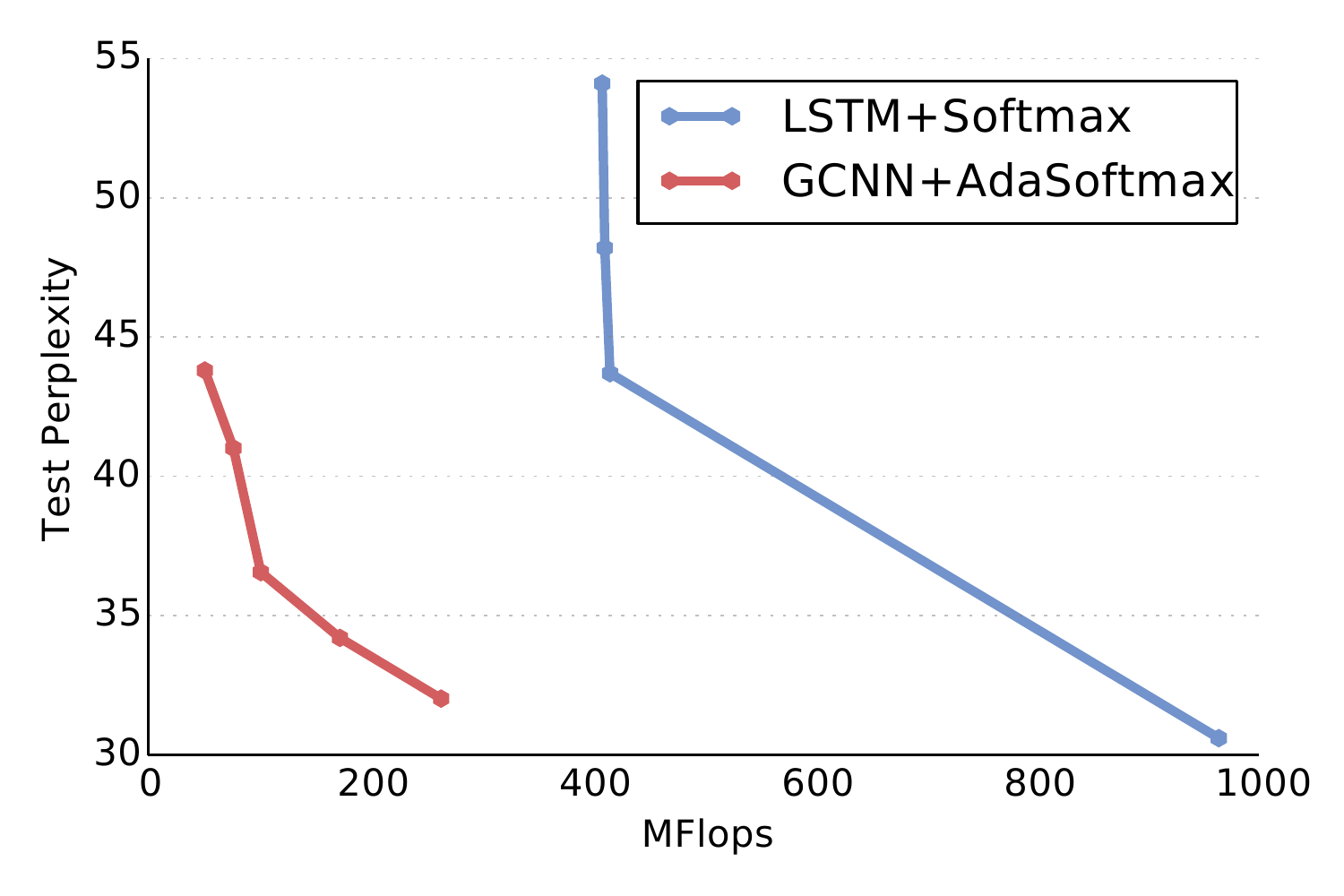}
 \caption{In comparison to the state-of-the-art \cite{jozefowicz2016exploring} which uses the full softmax, the adaptive softmax approximation greatly reduces the number of operations required to reach a given perplexity.}
 \label{fig:flops}
\end{figure}

We find the GCNN outperforms the comparable LSTM results on Google billion words. To accurately compare these approaches, we control for the same number of GPUs and the adaptive softmax output model \cite{2016arXiv160904309G}, as these variables have a significant influence on performance. In this setting, the GCNN reaches $38.1$ test perplexity while the comparable LSTM has $39.8$ perplexity (Table~\ref{fig:gbw}). 

Further, the GCNN obtains strong performance with much greater computational efficiency. Figure~\ref{fig:flops} shows that our approach closes the previously significant gap between models that use the full softmax and models with the usually less accurate hierarchical softmax. Thanks to the adaptive softmax, the GCNN only requires a fraction of the operations to reach the same perplexity values. The GCNN outperforms other single model state-of-the-art approaches except the much larger LSTM of \citet{jozefowicz2016exploring}, a model which requires more GPUs and the much more computationally expensive full softmax. In comparison, the largest model we have trained reaches $31.9$ test perplexity compared to the $30.6$ of that approach, but only requires training for 2 weeks on 8 GPUs compared to 3 weeks of training on 32 GPUs for the LSTM. Note that these results can be improved by either using mixtures of experts \citep{ShazeerMMDLHD17} or ensembles of these models. 



\begin{table}[ht]
  \centering \small
  \begin{tabular}{ l c r }
    \hline
    Model & Test PPL & Hardware\\
    \hline
        LSTM-1024 \cite{2016arXiv161204426G} & 48.7 & 1 GPU\\
        GCNN-8  & 44.9 & 1 GPU\\
        GCNN-14 & 37.2 & 4 GPUs\\
    \end{tabular}
   \caption{Results for single models on the WikiText-103 dataset.}
 \label{fig:wiki}
\end{table}
\footnotetext{appeared after submission}

Another relevant concern is if the GCNN's fixed context size can thoroughly model long sequences. 
On Google Billion Word, the average sentence length is quite short --- only 20 words. We evaluate on WikiText-103 to determine if the model can perform well on a dataset where much larger contexts are available. On WikiText-103, an input sequence is an entire Wikipedia article instead of an individual sentence - increasing the average length to 4000 words. However, the GCNN outperforms LSTMs on this problem as well (Table~\ref{fig:wiki}). The GCNN-8 model has 8 layers with $800$ units each and the LSTM has $1024$ units. 
These results show that GCNNs can model enough context to achieve strong results. 

We evaluated on the Gigaword dataset following \citet{Chen2016StrategiesFT} to compare with fully connected models. We found that the fully connected and convolutional network reach respectively 55.6 and 29.4 perplexity. We also ran preliminary experiments on the much smaller Penn tree bank dataset. When we score the sentences independently, the GCNN and LSTM have comparable test perplexity with 108.7 and 109.3 respectively. However, it is possible to achieve better results by conditioning on previous sentences. Unlike the LSTM, we found that the GCNN overfits on this quite small dataset and so we note the model is better suited to larger scale problems.

\subsection{Computational Efficiency}

\begin{table}[ht]
  \centering
  \begin{tabular}{ l c c c }
    \hline
     & \multicolumn{2}{c}{Throughput}  & Responsiveness  \\
     & (CPU) & (GPU) & (GPU) \\
    \hline
        LSTM-2048 & 169 & 45,622 & 2,282 \\
        GCNN-9 & 121 & 29,116 & 29,116 \\
        GCNN-8 {\small Bottleneck} & {\bf 179} & {\bf 45,878} & {\bf 45,878} \\
    \end{tabular}
   \caption{Processing speed in tokens/s at test time for an LSTM with 2048 units and GCNNs achieving 43.9 perplexity on Google Billion Word. The GCNN with bottlenecks improves the responsiveness by 20 times while maintaining high throughput.}
 \label{fig:speed}
\end{table}

\begin{figure*}[ht]
   \centering
   \includegraphics[width=.45\textwidth]{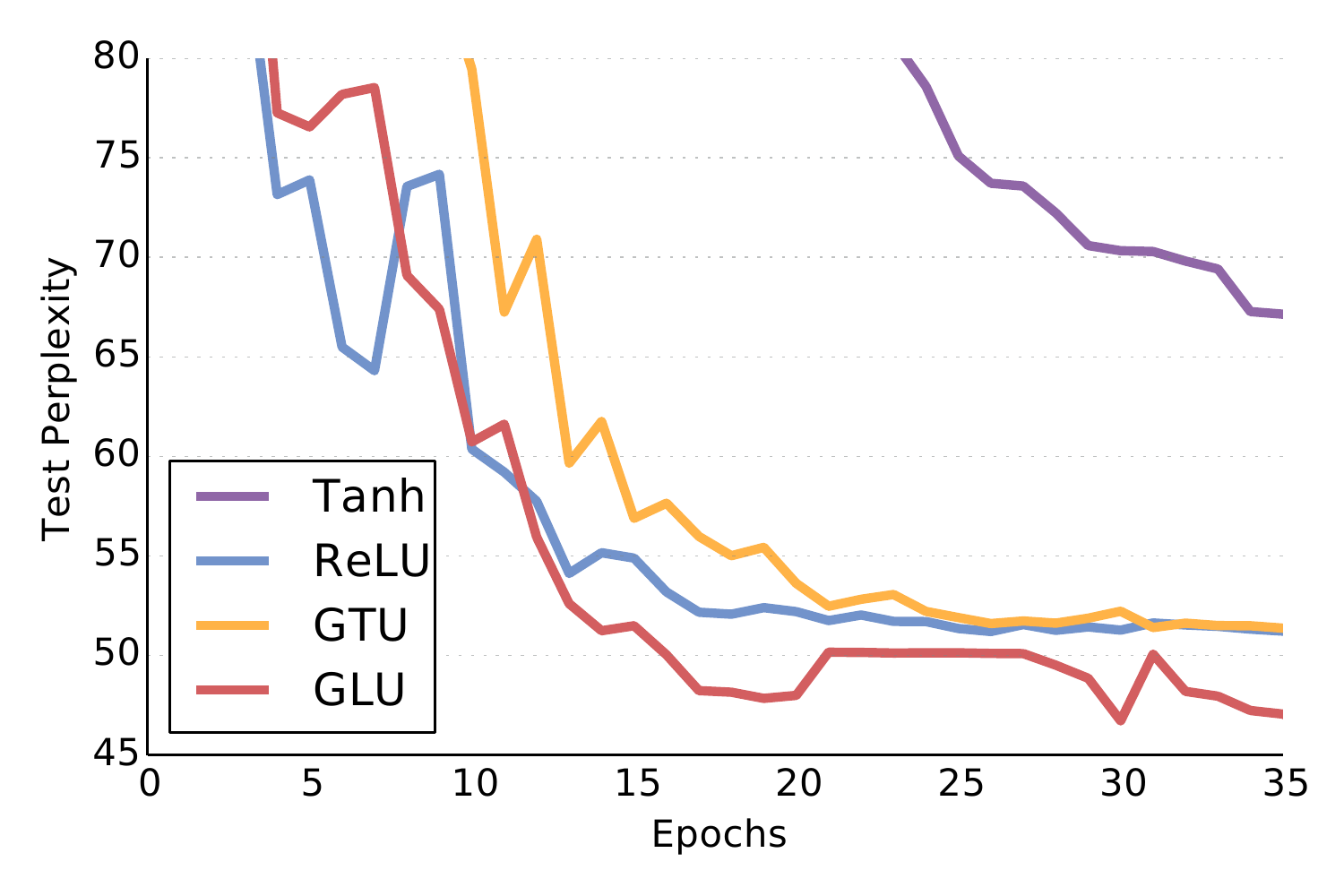}
   \includegraphics[width=.45\textwidth]{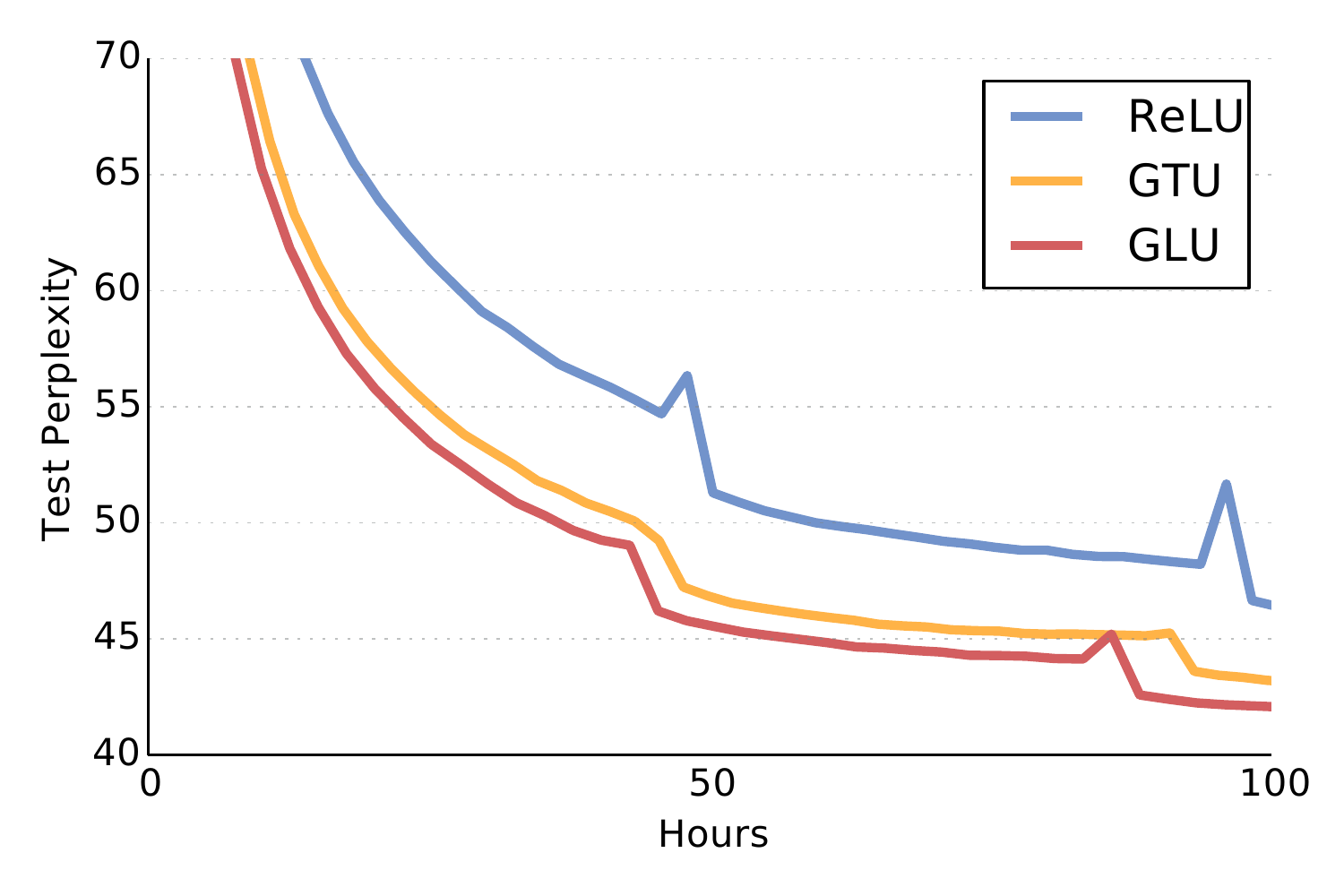}
 \caption{Learning curves on WikiText-103 (left) and Google Billion Word (right) for models with different activation mechanisms. Models with gated linear units (GLU) converge faster and to a lower perplexity.}
 \label{fig:gating}
\end{figure*}

Computational cost is an important consideration for language models.
Depending on the application, there are a number of metrics to consider.
We measure the \emph{throughput} of a model as the number of tokens that can be processed per second.
Throughput can be maximized by processing many sentences in parallel to amortize sequential operations.
In contrast, \emph{responsiveness} is the speed of processing the input sequentially, one token at a time.
Throughput is important because it indicates the time required to process a corpus of text and responsiveness is an indicator of the time to finish processing a sentence.
A model can have low responsiveness but high throughput by evaluating many sentences simultaneously through batching.
In this case, such a model is slow in finishing processing individual sentences, but can process many sentences at a good rate.

We evaluate the throughput and responsiveness for models that reach approximately $43.9$ perplexity on the Google Billion Word benchmark.
We consider the LSTM with $2048$ units in Table~\ref{fig:gbw}, a GCNN-8Bottleneck with 7 Resnet blocks that have a bottleneck structure as described by \citep{he2015deep} and a GCNN-8 without bottlenecks. A bottleneck block wedges a $k>1$ convolution between two $k=1$ layers. This designs reduces computational  cost by reducing and increasing dimensionality with the $k=1$ layers so that the convolution operates in a lower dimensional space. 
Our results show that the use of bottleneck blocks is important to maintaining computational efficiency.


The throughput of the LSTM is measured by using a large batch of $750$ sequences of length $20$, resulting in $15,000$ tokens per batch. The responsiveness is the average speed to process a sequence of $15,000$ contiguous tokens.
Table~\ref{fig:speed} shows that the throughput for the LSTM and the GCNN are similar.
The LSTM performs very well on GPU because the large batch size of $750$ enables high parallelization over different sentences.
This is because the LSTM implementation has been thoroughly optimized and uses cuDNN, whereas the cuDNN implementation of convolutions is not been optimized for the 1-D convolutions we use in our model.
We believe much better performance can be achieved by a more efficient 1-D cuDNN convolution.
Unlike the LSTM, the GCNN can be parallelized both over sequences as well as across the tokens of each sequence, allowing the GCNN to have 20x higher responsiveness. 


\begin{figure*}[ht]
   \centering
   \includegraphics[width=.45\textwidth]{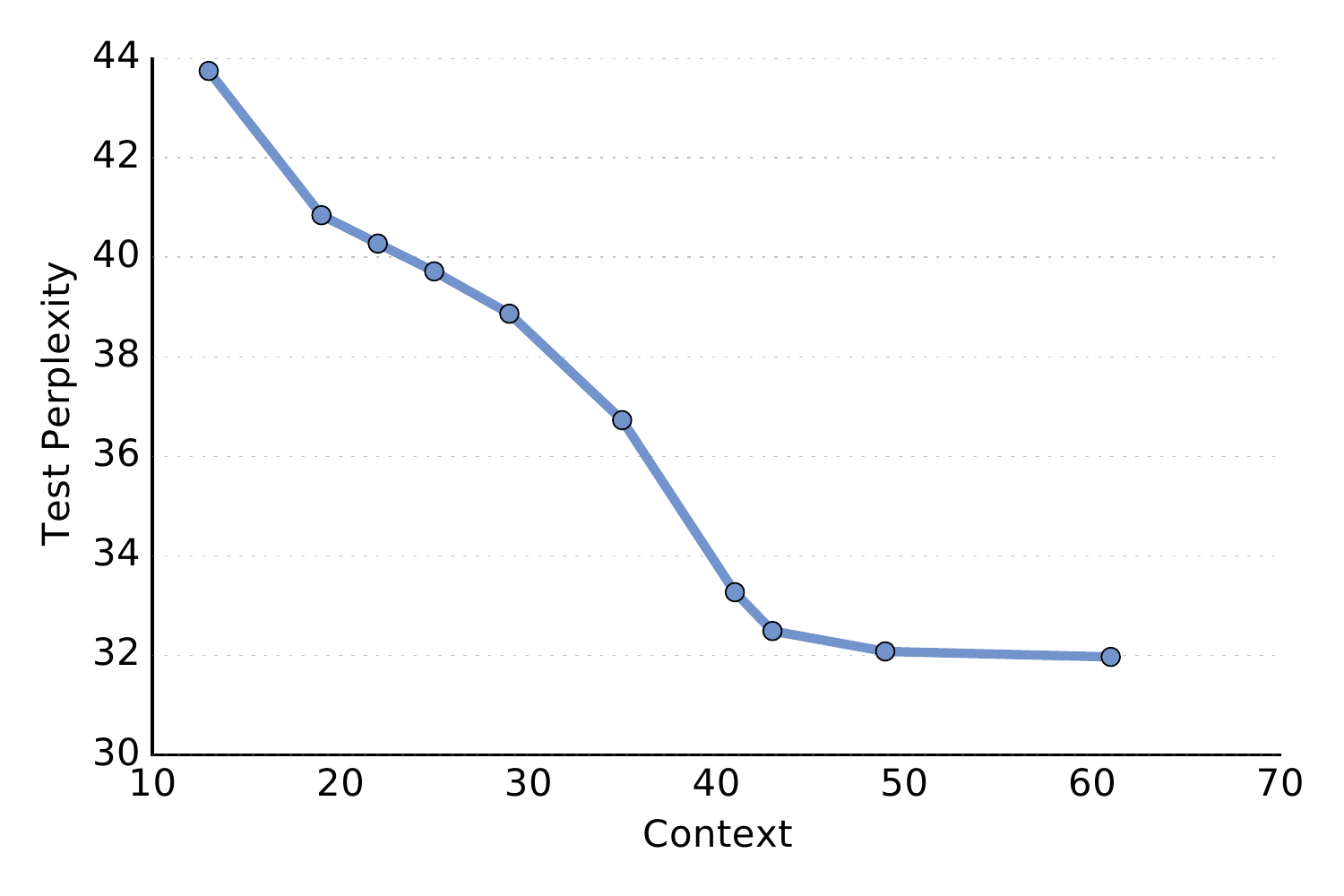}
   \includegraphics[width=.45\textwidth]{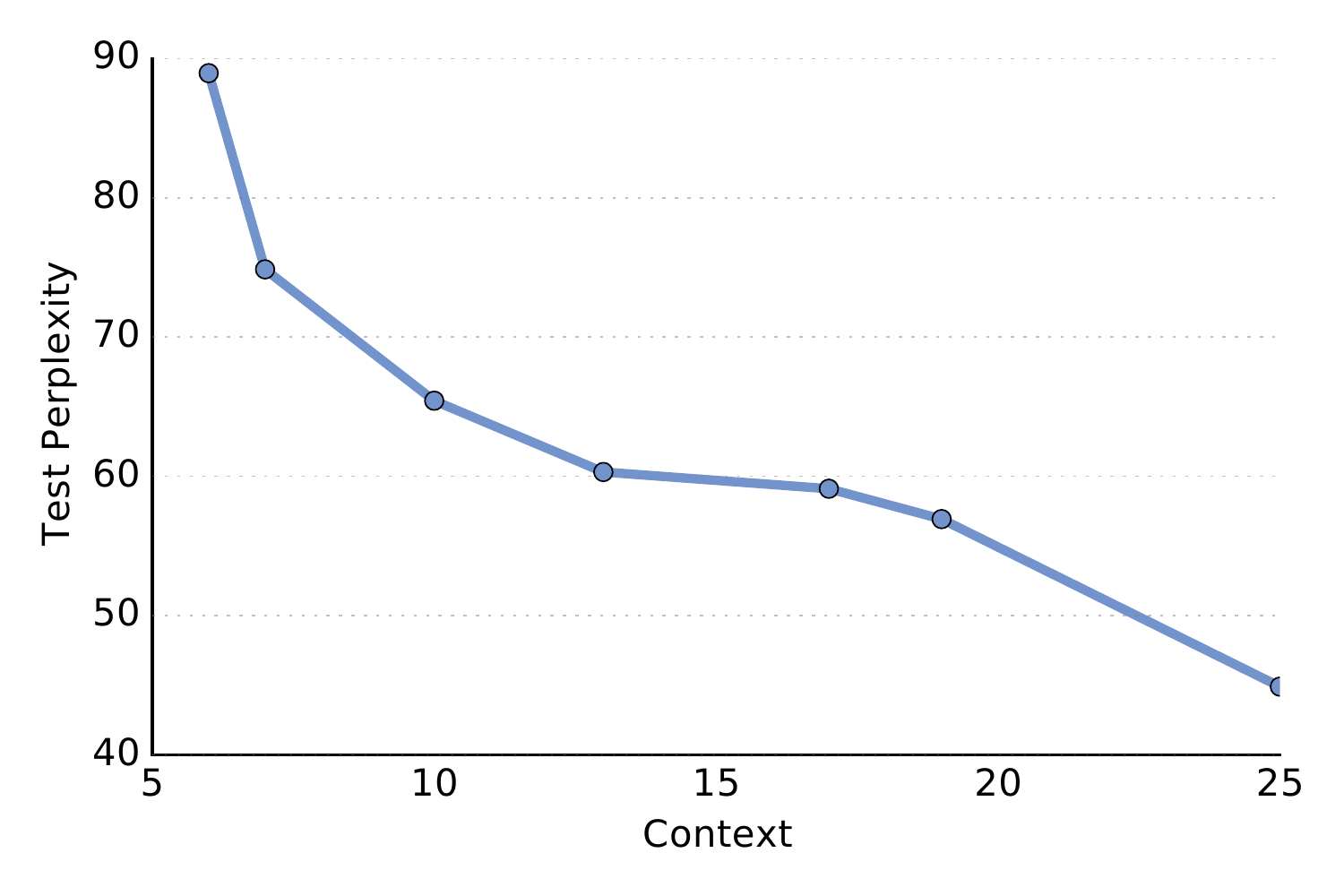}
 \caption{Test perplexity as a function of context for Google Billion Word (left) and Wiki-103 (right). We observe that models with \
bigger context achieve better results but the results start diminishing quickly after a context of 20.}
 \label{fig:context}
\end{figure*}

\subsection{Gating Mechanisms}\label{sec:gating}

In this section, we compare the gated linear unit with other mechanisms as well as to models without gating.
We consider the LSTM-style gating mechanism (GTU) $\text{tanh}({\bf X} \ast {\bf W} + {\bf b})\otimes \sigma({\bf X} \ast {\bf V} + {\bf c})$ of \citep{oord2016conditional} and networks that use regular ReLU or Tanh activations.
Gating units add parameters, so for fair comparison, we carefully cross-validate models with a comparable number of parameters.
Figure~\ref{fig:gating} (left) shows that GLU networks converge to a lower perplexity than the other approaches on WikiText-103.
Similar to gated linear units, the ReLU has a linear path that lets the gradients easily pass through the active units.
This translates to much faster convergence for both the ReLU and the GLU.
On the other hand, neither Tanh nor GTU have this linear path, and thus suffer from the vanishing gradient problem. In the GTU, both the inputs as well as the gating units can cut the gradient when the units saturate.

Comparing the GTU and Tanh models allows us to measure the effect of gating since the Tanh model can be thought of as a GTU network with the sigmoid gating units removed.
The results (Figure~\ref{fig:gating}, left) show that the gating units make a vast difference and provide useful modeling capabilities, as there is a large difference in the performance between GTU and Tanh units.
Similarly, while ReLU unit is not an exact ablation of the gating units in the GLU, it can be seen as a simplification $\text{ReLU}({\bf X}) = {\bf X}\otimes({\bf X} > 0)$ where the gates become active depending on the sign of the input. Also in this case, GLU units lead to lower perplexity.


In Figure~\ref{fig:gating} (right) we repeat the same experiment on the larger Google Billion Words dataset.
We consider a fixed time budget of $100$ hours because of the considerable training time required for this task.
Similar to WikiText-103, the gated linear units achieve the best results on this problem.
There is a gap of about 5 perplexity points between the GLU and ReLU which is similar to the difference between the LSTM and RNN models measured by \cite{jozefowicz2016exploring} on the same dataset.

\subsection{Non-linear Modeling}

The experiments so far have shown that the gated linear unit benefits from the linear path the unit provides compared to other non-linearities.
Next, we compare networks with GLUs to purely linear networks and networks with bilinear layers in order to measure the impact of the non-linear path provided by the gates of the GLU.
One motivation for this experiment is the success of linear models on many natural language processing tasks \cite{manning1999foundations}. We consider deep \emph{linear} convolutional networks where the layers lack the gating units of the GLU and take the form $h_l({\bf X}) = {\bf X} \ast {\bf W} + {\bf b}$.
Stacking several layers on top of each other is simply a factorization of the model which remains linear up to the softmax, at which point it becomes log-linear.
Another variation of GLUs are bilinear layers \citep{mnih2007three} which take the form $h_l({\bf X}) = ({\bf X} \ast {\bf W} + {\bf b})\otimes ({\bf X} \ast {\bf V} + {\bf c})$.

\begin{figure}[ht]
   \centering
   \includegraphics[width=.5\textwidth]{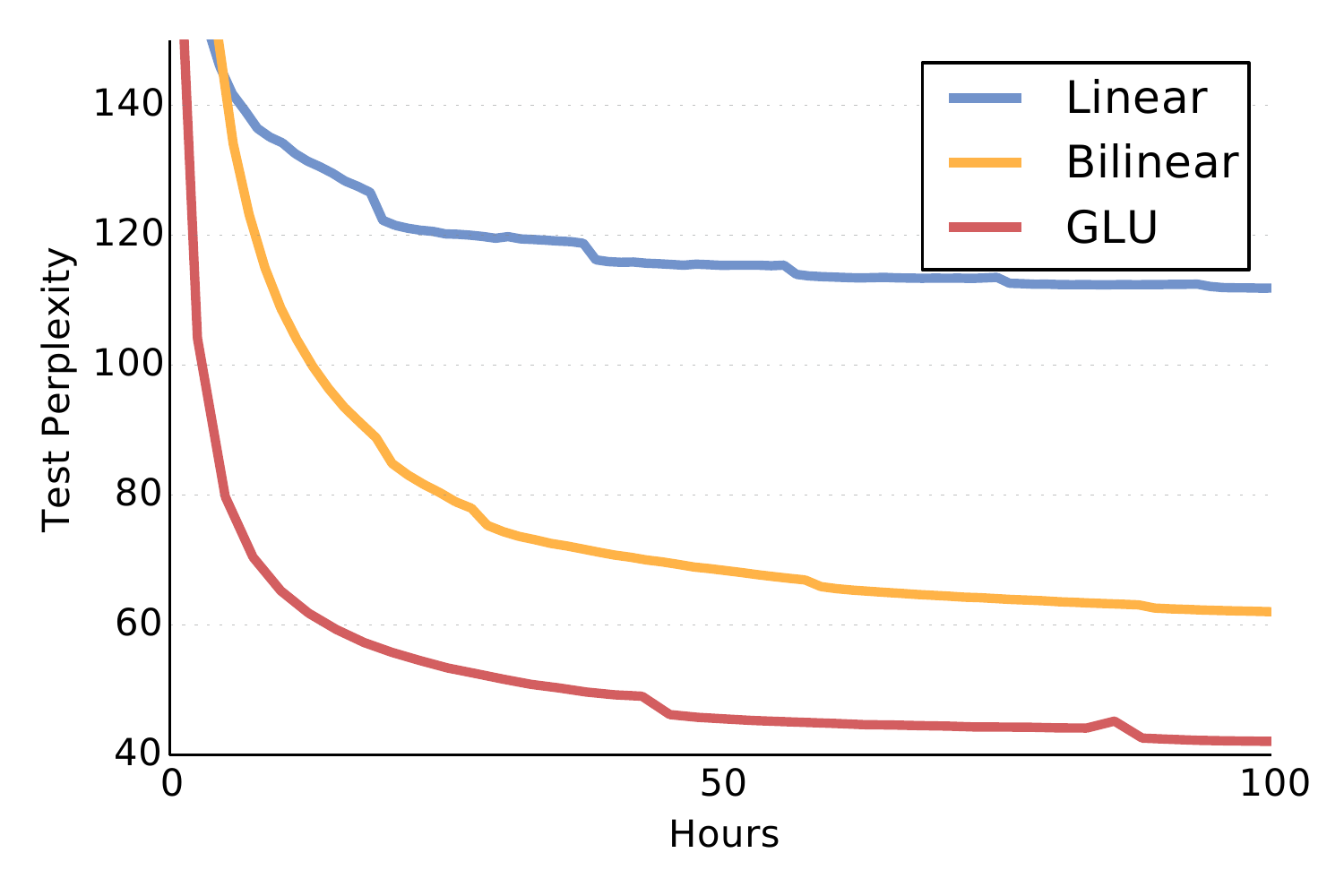}
 \caption{Learning curves on Google Billion Word for models with varying degrees of non-linearity.}
 \label{fig:nonlinear}
\end{figure}


Figure~\ref{fig:nonlinear} shows that GLUs perform best, followed by bilinear layers and then linear layers.
Bilinear layers improve over linear ones by more than $40$ perplexity points, and the GLU improves another $20$ perplexity points over the bilinear model.
The linear model performs very poorly at perplexity $115$ even compared to $67.6$ of a Kneser-Ney 5-gram model, even though the former has access to more context. Surprisingly, the introduction of the bilinear units is enough to reach $61$ perplexity on Google Billion Word, which surpasses both Kneser-Ney 5-gram models and the non-linear neural model of \cite{ji2015blackout}.




\subsection{Context Size}


Figure~\ref{fig:context} shows the impact of context size for the gated CNN. We tried different combinations of network depth and kernel widths for each context size and chose the best performing one for each size.
Generally, larger contexts improve accuracy but returns drastically diminish with windows larger than 40 words, even for WikiText-103 where we may condition on an entire Wikipedia article.
This means that the unlimited context offered by recurrent models is not strictly necessary for language modeling.
Furthermore, this finding is also congruent with the fact that good performance with recurrent networks can be obtained by truncating gradients after only 40 timesteps using truncated back propagation through time. Figure~\ref{fig:context} also shows that WikiText-103 benefits much more from larger context size than Google Billion Word as the performance degrades more sharply with smaller contexts.
WikiText-103 provides much more context than Google Billion Word where the average sentence size is 20.
However, while the average size of the documents is close to 4000 tokens, we find that strong performance can be achieved with a context size as low as 30 tokens.

\subsection{Training}

\begin{figure}[ht]
   \centering
   \includegraphics[width=.5\textwidth]{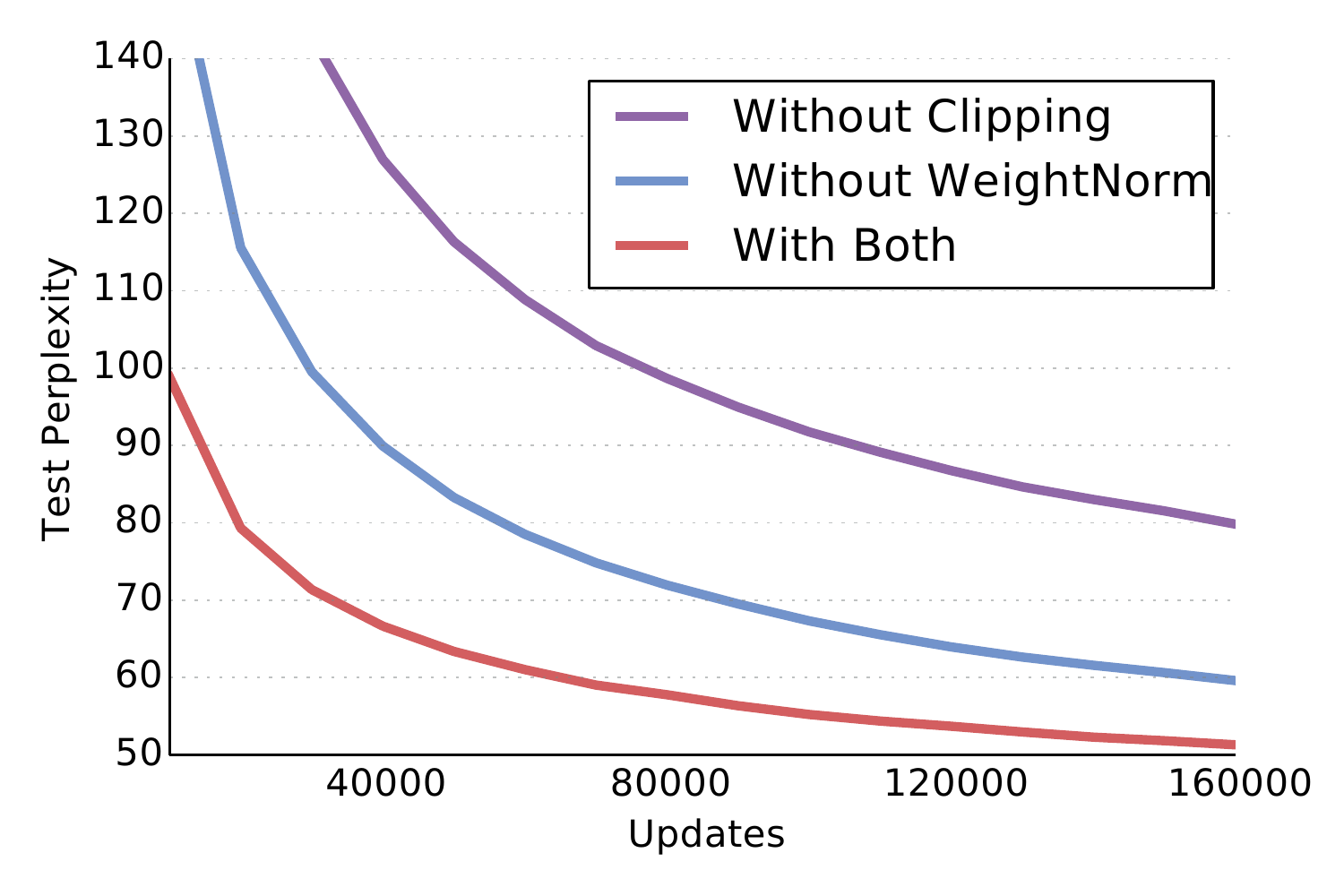}
 \caption{Effect of weight normalization and gradient clipping on Google Billion Word.}
 \label{fig:training}
\end{figure}

In this section, we perform an ablation study of the impact of weight normalization and gradient clipping.
We separately cross-validate the hyper-parameters of each configuration to make the comparison fair.
Due to the high cost of each of these experiments, we only consider a single iteration over the training data.
Figure~\ref{fig:training} shows that both methods significantly speed up convergence.
Weight normalization in particular improves the speed by over two times.
This speedup is partly due to the ability to use much larger learning rates ($1$ instead of $0.01$) than would otherwise be possible. Both clipping and weight normalization add computational overhead, but it is minor compared to the large gains in convergence speed.

\section{Conclusion}

We introduce a convolutional neural network for language modeling with a novel gating mechanism.
Compared to recurrent neural networks, our approach builds a hierarchical representation of the input words that makes it easier to capture long-range dependencies, similar in spirit to the tree-structured analysis of linguistic grammar formalisms.
The same property eases learning since features are passed through a fixed number of layers and non-linearities, unlike for recurrent networks where the number of processing steps differs depending on the position of the word in the input. The results show that our gated convolutional network achieves a new state of the art on WikiText-103.
On the Google Billion Word benchmark, we show competitive results can be achieved with significantly fewer resources.


\section*{Acknowledgments}

We would like to thank Ben Graham, Jonas Gehring, Edouard Grave, Armand Joulin and Ronan Collobert for helpful discussions.


\nocite{DBLP:journals/corr/WangLLJL15}

{\small
\bibliography{main}
\bibliographystyle{icml2016}
}

\end{document}